\documentclass[12pt]{article}

\usepackage{tikz}
\usepackage{fullpage}
\usepackage{amsmath}
\usepackage{setspace}
\usepackage{xspace}
\usepackage[ruled]{algorithm2e}
\usepackage{url}
\usepackage{authblk}
\usepackage{hyperref}

\def\cF{\mathcal{F}}

\def\cE{\mathcal{E}}
\def\cO{\mathcal{O}}

\def\inmessage{\alpha}
\def\outmessage{\rho}

\begin{document}

\onehalfspacing

\title{Clustering by transitive propagation}

\author{Vijay Kumar$^*$ and Dan Levy\footnote{Email: \url{vsreeniv@cshl.edu}, \url{levy@cshl.edu}. Both authors contributed equally to this work.}
}
\date{June 4, 2015}

\affil{
Simons Center for Quantitative Biology \\
Cold Spring Harbor Laboratory \\
1 Bungtown Road, \\
Cold Spring Harbor, NY 11724.
}

\maketitle

\begin{abstract}

We present a global optimization algorithm for clustering data given the ratio of likelihoods that each pair of data points is in the same cluster or in different clusters.
To define a clustering solution in terms of pairwise relationships, a necessary and sufficient condition is that belonging to the same cluster satisfies transitivity.
We define a global objective function based on pairwise likelihood ratios and a transitivity constraint over all triples, assigning an equal prior probability to all clustering solutions.
We maximize the objective function by implementing max-sum message passing on the corresponding factor graph to arrive at an $\cO(N^3)$ algorithm.
Lastly, we demonstrate an application inspired by mutational sequencing for decoding random binary words transmitted through a noisy channel.
\end{abstract}

\newpage

\section{Introduction}

Most algorithms for clustering data points determine clusters by minimizing in-cluster differences. 
In this paper, we consider the clustering problem wherein the data points are governed by two likelihood functions: $f_0(i, j)$ describing the probability that two data points $i$ and $j$ are from the same cluster, and $f_1(i, j)$ describing the probability that $i$ and $j$ derive from different clusters.
We use these two functions to assign a non-zero likelihood to any legal clustering configuration. 
This likelihood function is a product of $f_0$ and $f_1$ terms over all pairs of data-points.
We include with this likelihood a second term that constrains the  pair-wise assignments of ``same" or ``different" such that same-ness is transitive: a necessary and sufficient condition for ensuring a legal clustering configuration.
This constraint term, acting on all triples $(i, j, k)$, determines a uniform prior on the space of all distinct clustering solutions.

As in the case of affinity propagation \cite{frey2007clustering}, we first describe the factor graph \cite{kschischang2001factor} determined by our likelihood function, and use max-sum message passing \cite{pearl2014probabilistic} to identify a clustering configuration that maximizes the posterior distribution given our observed data points. 
The result is a clustering algorithm that is $\cO(N^3\times K)$ in complexity and $\cO(N^3)$ in memory usage, where $N$ is the number of data-points and $K$ is the number of iterations to convergence. In our experience, convergence is rapid and $K$ is typically very small.
The optimal clustering solution is a minimal energy configuration such that points are in the same cluster when they experience a net attractive force and  in different clusters when the net force is repulsive. 
This algorithm has the added benefit of not requiring an \textit{a priori} number of clusters.

In the next section, we calculate the posterior distribution whose maximization determines the optimal clustering. 
In section 3, we describe the factor graph for this distribution and describe our algorithm based on message passing. 
In section 4, we consider a detailed example that illustrates the method, and in section 5, we conclude with a summary of results, some trivial extensions, and future directions in applying relational constraints in factor graphs.

\section{Calculating the posterior distribution}

\subsection*{Notation}
Throughout this paper we will use the following notation. 
\begin{eqnarray*}
I &= & \{1, 2, \cdots, N\} \mbox{, the data points}\, , \\
E & = & \{(i,j)\ | \ i, j \in I, \, i \neq j, \ (i,j) \equiv (j,i)\} \, , 
\mbox{ the edges,}\\ 
T & = & \{(i,j,k)\ | \ (i,j) \in E, \ (j,k) \in E, \ (k,i) \in E\}\, , \mbox{ 
the triples,}\\
f_0 & = & P(i, j \ | \ i \ \textrm{and} \ j \  \textrm{are in the same 
cluster}) \\
f_1 & = & P(i, j \ | \ i \ \textrm{and} \ j \  \textrm{are in different clusters})
\end{eqnarray*}
We consider the fully connected graph $G$ with nodes $I$ and edges $E$. We 
assign a color to the edges of $G$ such that any edge is either blue = 0 or red = 1. The hypothesis matrix is a function $H: \, E \rightarrow \{0,1\}$, 
\begin{center}
$H_{ij} = $
\begin{tabular}{cc}
0,  &  $i,j$ belong to the same cluster (blue edge)\\
1,  &  $i,j$ belong to different clusters (red edge) 
\end{tabular}
\end{center}
For any hypothesis matrix we can compute the likelihood as
\begin{eqnarray}
L(I,H) = P(I | H ) = \prod_{(i,j) \in E} f_1(i, j)^{H_{ij}} f_0(i, j)^{1-H_{ij}}
\end{eqnarray}

\begin{figure}[t]
\includegraphics[width=0.75\textwidth]{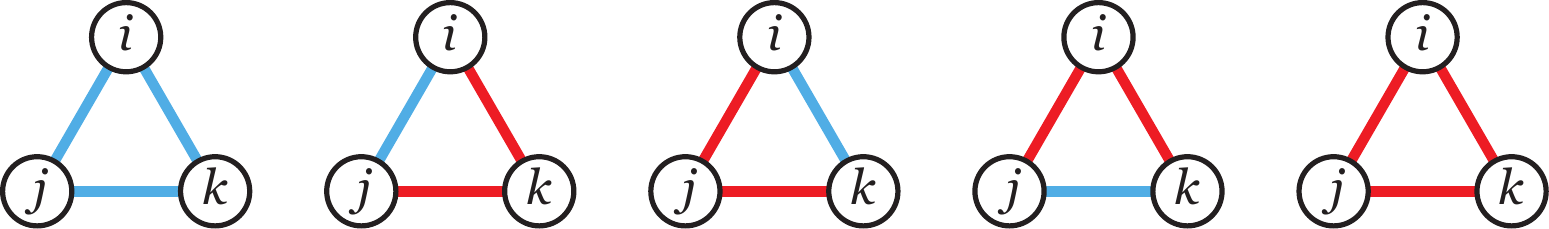}
\centering
\caption{Valid configurations of hypothesis. Blue edge = 0 and red edge = 1.}
\label{fig:valid-configurations}
\end{figure}

We assume that every clustering is equally likely, equivalent to a uniform 
prior over all $H$ obeying the transitivity condition,
\begin{equation}
P_{prior}(H) = \left\{\begin{array}{cll}
\frac{1}{B_N} & , & \mbox{ H represents a valid clustering} \\
0 & , &\ \mbox{ otherwise} 
\end{array}\right\}\, ,
\end{equation}
Here $B_N$ is the $N$-th Bell number that counts the total number of partitions of $N$ data points.
$H$ represents a valid clustering when every triple $(i,j,k) \in T$ satisfies the transitivity condition. The valid configurations for a single triple are shown in Figure \ref{fig:valid-configurations}. 
We can therefore express the uniform prior as a product over all triples:
\begin{eqnarray}
P_{prior}(H) & = & \frac{1}{B_N}\prod_{(i,j,k) \in T} V_{ijk}\, , \mbox{ where } 
\label{eq:simple-prior}\\
V_{ijk}  & = &
\left\{\begin{array}{cl}
1  & (H_{ij}, H_{jk}, H_{ki}) = (1,1,1), (1,1,0), (1, 0, 1), (0,1,1), (0,0,0) \\
0  & (H_{ij}, H_{jk}, H_{ki}) = (0,0,1), (0,1,0), (1,0,0) 
\end{array}\right. \label{eq:simple-prior2} \, .
\end{eqnarray}
For further details about the choice of prior and its consequences we refer the reader to Appendix \ref{sec:choice-of-prior}.

The posterior distribution over possible hypotheses $H$ can be calculated using 
the likelihood function and prior defined above
\begin{equation}
P(H | I) =  \frac{P(I|H) P_{prior}(H)}{\sum_H P(I|H) P_{prior}(H)} 
\label{eq:bayes-rule}\, .
\end{equation}
The sum in the denominator is the sum over all $2^{N(N-1)/2}$ possible $H$. The 
prior restricts the posterior distribution to valid clustering solutions. 
We define the optimal clustering as the hypothesis matrix $H^*$ that maximizes 
the posterior probability,
\begin{eqnarray}
H^* 	& = & \mathop{\mathrm{argmax}}_{H} \ P(H | I)\, . \\
		& = & \mathop{\mathrm{argmax}}_{H}  \big[ \log P(I | H) + 	\log 
		P_{prior}(H)   \big] \\
		& = & \mathop{\mathrm{argmax}}_{H} \left[ \sum_{(i,j) \in E} H_{ij} 
		\log \frac{f_1(i, j)}{f_0(i, j)} + \sum_{(i,j,k) \in T} \log V_{ijk} 
		(H_{ij}, H_{jk}, H_{ki}) \right]\, .
\end{eqnarray}
In arriving at the final result we have dropped terms that are independent of 
$H_{ij}$ since they do not effect the result of the argmax operation.

To simplify notation we define an objective function
\begin{equation}
\boxed{\cF(H) = \sum_{(i,j)\in E} S_{ij}(H_{ij}) + \sum_{(i,j,k) \in T} 
\delta_{ijk} (H_{ij}, H_{jk}, H_{ik})\, ,} \label{eq:objective-function}
\end{equation}
where $S_{ij}(H_{ij}) := H_{ij} \log \frac{f_1(i, j)}{f_0(i, j)} $ and 
$\delta_{ijk} := \log V_{ijk}$. 

\subsubsection*{Interpretation in terms of energy minimization}

We can define a Hamiltonian or an energy function, 
\begin{eqnarray}
\cE = -\sum_{(i,j) \in E}  \left[H_{ij}\log f_1(i, j)  + (1-H_{ij}) \log f_0(i, j) 
\right]  - \sum_{(i,j,k) \in T} \delta_{ijk} (H_{ij}, H_{jk}, H_{ki}) 
\end{eqnarray}
over the space of all matrices $H$. Note that $\cE = -\cF + \mbox{constant}$. The optimal clustering is defined as the minimum of this energy function.
The terms $\log f_0$ and $\log f_1$ can be viewed as forces of attraction and repulsion.
For a given pair of points $i$, $j$, if $f_0 > f_1$ then the energy is lowered if $H_{ij} = 0$ or they are in the same cluster, and if $f_1 > f_0$ the energy is lowered when $H_{ij}=1$. 
In the absence of the prior term, the energy is minimized by the following solution
\begin{eqnarray}
H_{ij}^{\mbox{\scriptsize no prior}} = \left\{\begin{array}{ll} 
0, & f_0(i,j) > f_1(i,j) \\
1, & \mbox{otherwise} \end{array} \right\}\, .
\end{eqnarray}
This solution is applicable when the data point clusters are well separated. Moreover, we have constructed this optimal solution through independent decisions for every edge. 
The prior complicates the problem and introduces a three-point long-range interaction term that is infinitely repulsive when the transitivity condition is disobeyed. However, if $H_{ij}^{\mbox{\scriptsize no prior}}$ is consistent with transitivity, then it minimizes the energy and no further work is needed to identify an optimal configuration.

In the next section, we represent the objective function $\cF$ as a factor 
graph and use message passing to determine the configuration that maximizes the objective function.

\section{Maximizing the objective function}
\label{sec:message-passing}

\begin{figure}[t]
\centering
\includegraphics[width=4in]{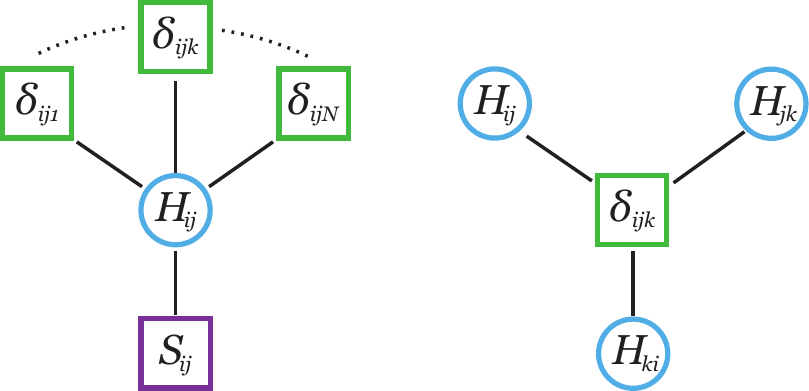}
\caption{The factor graph for the objective function $\cF$ defined in equation \eqref{eq:objective-function} is composed of two types of junctions. On the left is the sub-graph of the neighbors of the variable node $H_{ij}$ and on the right, the sub-graph of neighbors of the function node $\delta_{ijk}$.}
\label{fig:factor_graph}
\end{figure}

We can represent the objective function and its dependence on the hypothesis matrix $H$ with a \emph{factor graph} \cite{kschischang2001factor}. 
The factor graph consists of two types of nodes: \textit{variable nodes}, represented by a circle, for every independent hypothesis variable in $H$, and \textit{function nodes}, represented by a square for each summand in the objective function \eqref{eq:objective-function}.
When a function node $g$ depends on a variable $x$, we connect the nodes by an edge. 
Every variable node $H_{ij}$ has $(N-2)$ edges that connect it to function nodes $\delta_{ijk}$ for all $k \neq i, j$; every function node $\delta_{ijk}$ is connected to three variable nodes $H_{ij}$, $H_{jk}$ and 
$H_{ki}$; the function node $S_{ij}$ has only one edge to the $H_{ij}$ variable node. 
The factor graph is depicted in Figure \ref{fig:factor_graph}.

\begin{figure}[t]
\centering
\includegraphics[width=4in]{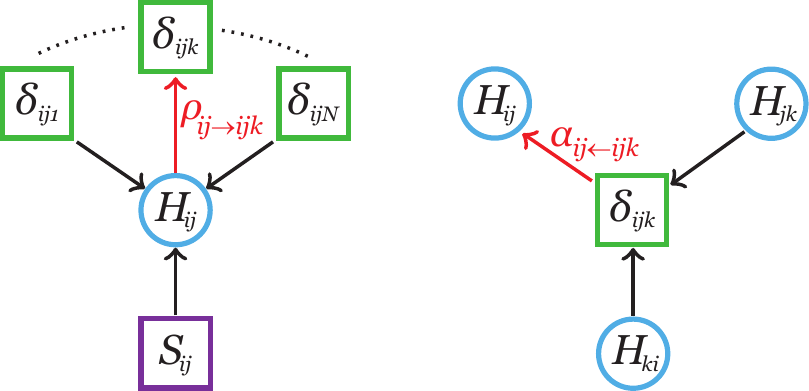}
\caption{As in Figure \ref{fig:factor_graph}, we show the transmitted and received messages for $H_{ij}$.}
\label{fig:messages}
\end{figure}

We use message passing on the factor graph to solve for $H^* = 
\mathop{\mathrm{argmax}}_{H} \cF$. This technique has been applied to a variety of problems in different fields as discussed in \cite{mezard2003passing}.
Since the factor graph has cycles, our approach is an example of loopy belief propagation \cite{pearl2014probabilistic}. 
The success of this method has been explained in terms of the accuracy of the Bethe free energy approximation \cite{yedidia2003understanding}.
Every message is a two-tuple as every hypothesis variable has two possible 
values. We denote the message transmitted from $H_{ij}$ to $\delta_{ijk}$ by 
$\outmessage_{ij \rightarrow ijk}$ and the received message by $\inmessage_{ij 
\leftarrow ijk}$ as shown in Figure \ref{fig:messages}. Both messages are functions of the corresponding variable 
node $H_{ij}$. The function node $S_{ij}$ continuously transmits the same 
message to $H_{ij}$.

The messages are updated as follows, first the variable nodes transmit to  
function nodes
\begin{eqnarray}
\outmessage_{ij\rightarrow ijk}(H_{ij}) = S_{ij}(H_{ij}) + \sum_{l \neq i, j, 
k} \alpha_{ij \leftarrow ijl}(H_{ij})\, ,
\end{eqnarray}
and then receive responses
\begin{eqnarray}
\alpha_{ij \leftarrow ijk}(H_{ij}) = \max_{H_{jk}, H_{ki}} \big[ 
\delta_{ijk}(H_{ij}, H_{jk}, H_{ki}) + \outmessage_{jk\rightarrow ijk}(H_{jk}) 
+ \outmessage_{ki \rightarrow ijk}(H_{ki})\big]\, .
\end{eqnarray}

This sequence of transmission and reception defines one iteration of the 
algorithm. At the end of each iteration, the configuration $H^*$ is given by
\begin{eqnarray}
H^*_{ij} = \mathop{\mathrm{argmax}}_{x = \{0,1\}} \left( S_{ij}(x) + \sum_{k 
\neq i,j} \alpha_{ij \leftarrow ijk}(x)\right) \label{eq:best-hypothesis}
\end{eqnarray}
We repeat, iterating through transmissions and receptions until $H^*$ is 
unchanged.

The message update rules can be considerably simplified. First, the $\outmessage_{ij\rightarrow ijk}$ messages can be eliminated, such that we need only compute updates for $\alpha_{ij \leftarrow ijk}$. Second, the solution $H_{ij}^*$ only depends on the combination $A_{ijk} := \alpha_{ij \leftarrow ijk}(1) - \alpha_{ij \leftarrow ijk}(0)$ so we do not need to calculate values for both states (blue and red) but only for the difference. Lastly, we introduce the auxiliary matrix $B_{ij} := \Delta S_{ij} + \sum_{k \neq i,j} 
A_{ijk}$ that reduces the complexity of the update procedure from $\cO(N^4)$ to $\cO(N^3)$. We refer the 
interested reader to the discussion in Appendix \ref{sec:simplify-update} for 
details. Here, we show the result in the form of an explicit algorithm that we call \hyperref[algo]{\emph{Transitive Propagation}}, 
which has complexity $\cO(N^3 \times K)$ and $\cO(N^3)$ memory usage, where $K$ 
is the (typically small) number of iterations to convergence.

\newpage
\begin{algorithm}[h]
\caption{Transitive Propagation}
\label{algo}

 \SetKwRepeat{Do}{do}{while}
 \KwData{$N$ data-points with distributions $f_{0,1}(i, j)$ for all $i, j \in 
 I$ and $\lambda \in (0,1)$.}
 \KwResult{Optimal hypothesis matrix $H^*$.}
Calculate $N \times N$ matrix $\Delta S_{ij} = \log f_1(i, j) - \log f_0(i, 
j)$.\\
Initialize $N\times N\times N$ matrix $A_{ijk} := 0$\;
Define convergence goal $M = 1000$\;
 \Do{$0 < m < M$}{
Compute $N\times N$ matrix $B$ : $B_{ij} = \Delta S_{ij} + \sum_{k \neq i,j} 
A_{ijk}$	\;
Compute update $\Delta A_{ijk}$ defined by 
	\noindent
	\begin{flalign*}
 \Delta A_{ijk} =  \max \left\{ 0, \ \ B_{jk} - A_{jki}+  B_{ki}  -  A_{kij} 
 \right\} 
	   -  \max \left\{ 0, \ \ B_{jk} - A_{jki},  \ \ B_{ki} - A_{kij} \right\} 
	   \, ;  
	\end{flalign*}
Perform update including the dampening factor $\lambda$:
	\noindent
	\begin{flalign*}
	A_{ijk}  \leftarrow & (1-\lambda)\ A_{ijk} + \lambda \  \Delta A_{ijk} \, ;
	\end{flalign*}
Calculate $\Delta B_{ij} = \sum_{k\neq i,j} \Delta A_{ijk}$ and 
\begin{eqnarray*}
m = -\mathop{\mathrm{min}}_{(i,j)\in E}\ \  \frac{B_{ij}}{\Delta 
B_{ij}}\, .
\end{eqnarray*}
 }
 Compute $N\times N$ matrix $H^*$: $H^*_{ij} = \left\{\begin{array}{ll} 1, & 
 B_{ij} \geq 0 \\ 0, & B_{ij} < 0 \end{array} \right. $
\end{algorithm}

\noindent
{\bf Convergence and dampening}

We have introduced a dampening factor $\lambda$ that helps the algorithm 
converge to a fixed point rather than a cycle. Small values of $\lambda$ 
promote convergence but also increase the running time of the algorithm. 
We find that the choice $\lambda = 0.5$ is a good balance between time to 
convergence and avoiding cycles. 

The entries in $A_{ijk}$ do not converge to fixed values, and this is to be 
expected because we do not normalize the messages after each iteration. 
The solution $\{H^*_{ij}\}$ only depends on the sign of the $B_{ij}$ matrix. 
Consequently, our convergence criterion is as follows: at each iteration we 
estimate the minimum number of iterations, $m$, it would take to change the 
sign of one entry in $B_{ij}$ and stop when the number of iterations reaches a 
defined threshold, $M$.

\section{Example: clustering random bit patterns}
In this section, we present the clustering problem that inspired the  development of transitive propagation.
Recently, one of the authors proposed a method to uniquely tag DNA molecules 
through a process of random mutagenesis. 
By marking each template molecule with a random pattern, we can resolve two 
difficulties that continue to plague high-throughput short-read sequencing: (1) 
counting DNA molecules accurately and (2) assembling DNA sequences across 
repeat regions that exceed a read length.
We do not discuss the details here, but refer the reader to the original paper 
\cite{levy2014facilitated}. 

The example we address in this section is an abstracted version of the first 
problem, known in the literature as the $K$-populations problem, and has been shown to be NP-hard \cite{parida2000partitioning}.
Assume we have $K$ initial copies of a DNA sequence containing $L$ mutable 
positions.
Our mutation protocol randomly assigns one of two letters with equal 
probability at each position, generating $K$ binary words of length $L$.
These templates are copied many times and a machine analyzes those copies, 
outputting a \textit{read} that matches the initial template's binary word but 
introduces errors at a rate of $p_e \ll 1$ per bit.
Starting with $N$ reads generated through this process, we would like to 
determine the number of initial templates $K$, assigning the reads to clusters 
that correspond to the same initial template.

We work in a regime where $N \gg K$ so that all templates are sampled and read 
by the sequencer. 
Since the error rate is low, we expect that the $N$ reads form $K$ 
clusters, where $K$ is the unknown number of templates that we wish to 
determine.
We begin by measuring the hamming distance $d_{ij}$ between all reads $i$ and 
$j$.
When two reads are in the same cluster, 
\begin{equation}
f_0(i, j) = \binom{L}{d_{ij}}x^{d_{ij}} ( 1-x)^{L-d_{ij}}\, 
,\mbox{ where } x = 2p_e(1-p_e)
\end{equation}
and when they belong to different clusters, 
\begin{equation}
f_1(i, j) = \binom{L}{d_{ij}}\frac{1}{2^L}\, .
\end{equation}

We generated $K$ templates of length $L=30$ bits and generated $N=10K$ reads by uniform sampling. We introduced errors at a rate of $p_e$ per bit. The results that we present were obtained by averaging over 100 simulations for various values of $K$ and $p_e$.

\begin{figure}[t]
\centering
\includegraphics[width=\textwidth]{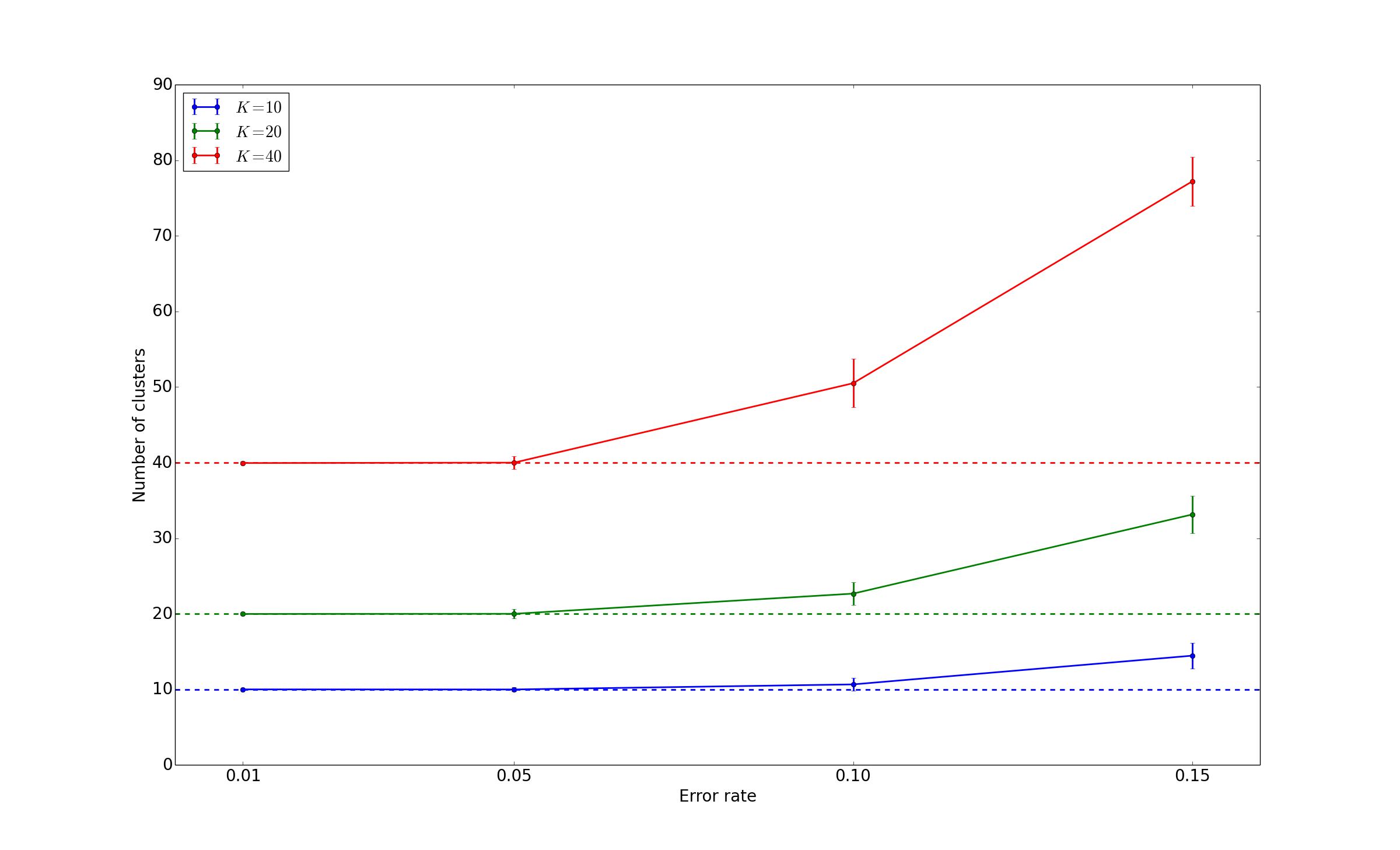}
\caption{The number of clusters obtained as a function of the error rate for different template counts, which are shown as dashed lines. The results are averaged over 100 simulations and the error bars denote one standard deviation. }
\label{fig:template-count}
\end{figure}
We performed computer simulations to evaluate our algorithm. 
We generated $K$ random templates of length $L=30$ bits for $K = 10, 20, 40$. We simulated $N = 10K$ reads with various error rates of $p_e = 0.01, 0.05, 0.10, 0.15$ per bit. 
Figure \ref{fig:template-count} shows the accuracy in determination of the template count as a function of the error rate averaged over 100 simulations.
We see accurate recovery of the template count even at high error rates of $p_e = 0.05$. 

\begin{figure}[t]
\includegraphics[width=7in]{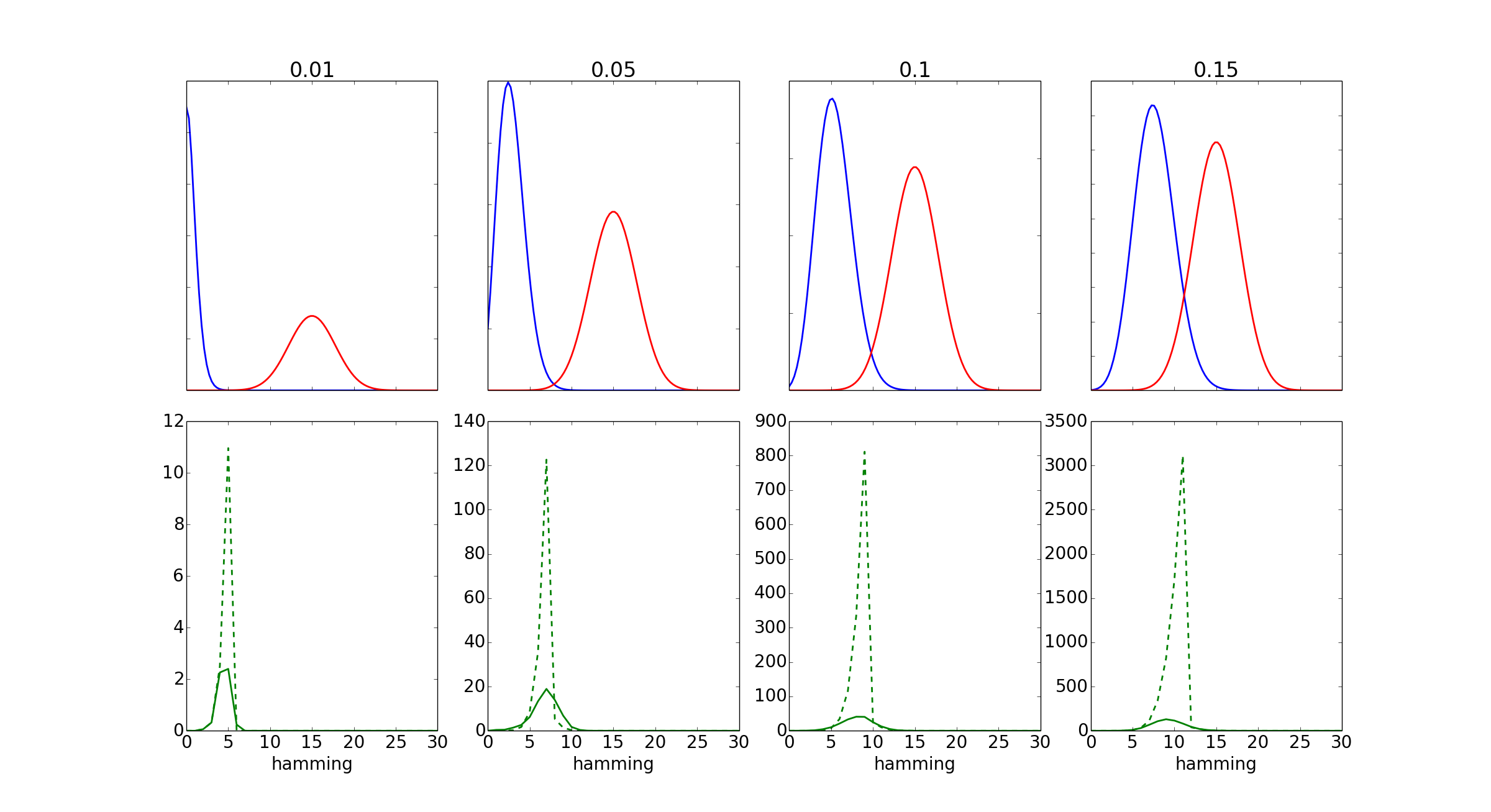}
\caption{The panel above shows the distributions $f_0$ (in blue) and $f_1$ (in red) for various values of the error rate $p_e$. In the corresponding panel below we show the average number of incorrect calls made by our algorithm (green, solid line) and by classifying each edge $i,j$ based on the likelihood ratio $f_1(i,j)/f_0(i,j)$ alone.}
\label{fig:incorrect-edges}
\end{figure}

Our algorithm is also very accurate in determining the correct clustering configuration when the error rate is high.
We fixed $K=50$ templates of length $L=30$ bits and generated $N=250$ reads for various values of error rate $p_e = 0.01, 0.05, 0.10, 0.20$ and performed 100 simulations. 
Our measure of accuracy is the number of edges that are mis-classified by the algorithm averaged over all the simulations. We plot the number of incorrect edges as a function of the hamming distance between the reads in Figure \ref{fig:incorrect-edges}.
As a reference, we also plot the number of incorrect edges if we classified each edge $i,j$ as red or blue based only on the likelihood ratio $f_1(i,j)/f_0(i,j)$. 
As expected, edges with very low or very high hamming distance are correctly inferred using both methods. For edges in the intermediate regime our method makes better inferences due to the transitive property.

\section{Discussion}
Transitive propagation is a useful algorithm for clustering data modeled by a balance of attractive and repulsive factors.
By imposing a naive prior, the method uniformly explores the space of all partitions of the data-points, enforcing no \textit{a prior} number of clusters or arbitrary similarity cut-off as required by other methods.
As described in Appendix \ref{sec:choice-of-prior}, the naive prior does impose a non-uniform probability on the number of clusters. However, even this prior distribution may be tuned.

\begin{figure}[t]
\centering
\includegraphics[width=4in]{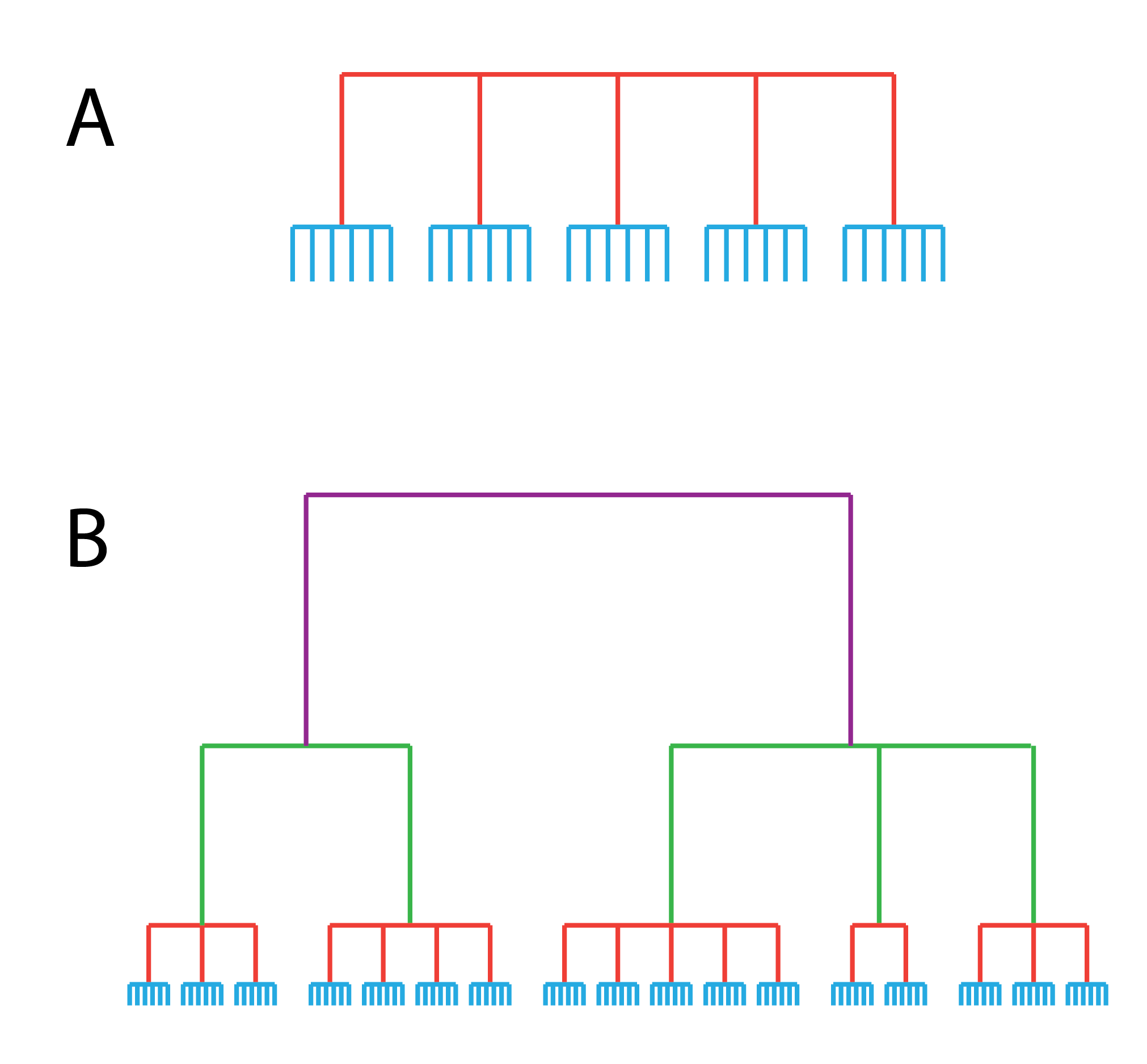}
\caption{Panel A shows an ultrametric tree corresponding to the transitivity propagation algorithm. In panel B, we extend the ultrametric property to include four distinct levels, each with a likelihood function $f_\alpha(i, j)$ for $\alpha = 0,1,2,3$ corresponding to multi-scale clustering.}
\label{fig:ultrametric}
\end{figure}

The transitive propagation algorithm can be extended in the following ways.
First, the existing algorithm implements max-sum message passing to identify a single configuration that maximizes the likelihood.
However, we can also implement sum-product message passing  to determine the marginal posterior probabilities that two data-points derive from a common cluster.
Such an algorithm would allow the selection of only the most confident edges, so as to discard outlying data-points.
Second, the existing framework assumes that the $f$ function depends on $i$ and $j$ such that all clusters follow the same distribution.
This limitation can be overcome through the inclusion of node-specific clustering parameters that enable variation in the intra-cluster distributions.

The methodology used in this paper to address clustering can be extended to other problems that we leave to future work.
First, the transitive constraint may be considered as the first non-trivial example of an integer valued ultra-metric which assumes one of two values: 0 or 1.
In this formulation, the prior constraint on $V_{ijk}$ in equation \ref{eq:simple-prior2} is identical to the ultra-metric property.
We can extend to higher order clusters by allowing a family of likelihood functions, $f_\alpha$ for $\alpha=0,1,...,L$, that measure increasingly divergent relationships between nodes (see figure \ref{fig:ultrametric}) and allow $H_{ij}$ to assume values of {0, 1, ... $L$}.
This modified algorithm enables multi-scale clustering.
Second, we can apply the same framework of constrained optimization to enforce relationships other than equality. 
For example, we may have data-points that obey a partial ordering.
The same constraints apply to $H_{ij}$ as before, however it is no longer the case that $H_{ij} = H_{ji}$. Depending on the nature of the data, the optimization function may depend on the four possible states for the pair ($H_{ij}, H_{ji}$) equivalent to the four possible cases: (1) $i$ and $j$ are the coincident, (2) $i$ precedes $j$, (3) $j$ precedes $i$, or (4) there is no relation between $i$ and $j$.

\subsection*{Acknowledgements}

We thank
Robert Aboukhalil,
Arjun Bansal,
Sharat Chikkerur,
Vishaka Datta,
Sarah Harris,
Ivan Iossifov,
Jude Kendall,
Bud Mishra,
Swagatam Mukhopadhyay,
Adam Siepel,
Vinay Satish,
Michael Schatz,
Michael Wigler, 
Boris Yamrom,
and the participants of QB Tea on May 13, 2015 for discussions, questions, and feedback that helped develop our ideas.
VK and DL are funded by CSHL grant 125217/QB-SIMONS.
This work was also supported by a grant from the Simons Foundation (SFARI award number 235988).

\bibliography{clustering_refs}{}
\bibliographystyle{ieeetr}

\appendix
\section{Some observations about our choice of prior}
\label{sec:choice-of-prior}
\def\cZ{\mathcal{Z}}
\def\cH{\mathcal{H}}

We can construct a family of conjugate priors for the problem of clustering $N$ data points parameterized by a real $N\times N$ matrix $X = [X_{ij}]$,
\begin{eqnarray}
P_{conj}(H \  |\  X, N) = \frac{1}{\cZ(X,N)}\, \prod_{(i,j) \in E} X_{ij}^{H_{ij}} 
\times 
\prod_{(i,j,k) \in T} V_{ijk}\, , \label{eq:conjugate-prior}
\end{eqnarray}
where $\cZ$ is a normalization factor. With this choice of prior, the posterior distribution in equation \eqref{eq:bayes-rule} becomes
\begin{eqnarray}
P(H \ | \ I, X, N) = P_{conj} (H \ | \ X', N)\, ,
\end{eqnarray}
where $X'_{ij} = X_{ij} + \log f_1(i,j) - \log f_0(i,j)$. In this section, we study a one-parameter sub-family given by $X_{ij} = x$, where $x$ is a non-negative real number,
\begin{eqnarray}
F(H, x, N) = \frac{1}{Z(x,N)}\, \prod_{(i,j) \in E} x^{H_{ij}} 
\times 
\prod_{(i,j,k) \in T} V_{ijk}\, . \label{eq:full-prior}
\end{eqnarray}
The uniform prior introduced in \eqref{eq:simple-prior} is a member of this one-parameter family, $P_{prior}(H) = F(H, x=1, N)$. The function $Z(x,N)$ is the overall normalization and is usually called the partition function
\begin{eqnarray}
Z(x,N) = \sum_{H} e^{- \cH(H, x,N)}\, 
\end{eqnarray}
where the Hamiltonian $\cH(H, x,N) =  \sum_{(i,j,k) \in T} 
\delta_{ijk} (H_{ij}, H_{jk}, H_{ik}) +  \sum_{(i,j)\in E} H_{ij} \log x$ is an example of a spin Hamiltonian where the $H_{ij}$ can be viewed as ``spin'' degrees of freedom and the parameter $\log x$ is the applied magnetic field. However, rather than the usual pairwise spin-spin interaction we have a 3-spin $\delta_{ijk}$ term. We study the phase diagram of this Hamiltonian as a function of $x$ and find an order-disorder transition at a critical value of $x = x_{critical}$ where $x_{critical} - 1 \sim \cO(\frac{\log N}{N})$. We suspect that this system has been studied in the vast literature on spin Hamiltonians and spin glasses but we are not aware of this.

Alternatively, it can be written as a sum over configurations satisfying the transitivity constraint
\begin{eqnarray}
Z(x,N) = \sum_{C} x^{b(C)}\, ,
\end{eqnarray}
where $C$ runs over all possible partitions of $N$ data points into clusters and $b(C)$ is the number of blue edges.
When $x=1$ we 
recover the uniform prior \eqref{eq:simple-prior}; and $Z(x=1, N) = B_N$ where 
$B_N$ are the Bell numbers that enumerate the total number of partitions of a 
set of $N$ elements. The limiting behavior is
\begin{eqnarray}
\lim_{x \rightarrow 0} Z(x,N) = 1\, , \ \lim_{x \rightarrow \infty} \frac{Z(x,N)}{x^{N(N-1)/2}} = 1\, .
\end{eqnarray}
The partition function $Z(x,N)$ satisfies a recurrence relation
\begin{gather}
Z(x,N+1) = \sum_{k=0}^N \binom{N}{k}\ x^{k(k+1)/2}\ Z(x,N-k)\, , \\
 Z(x,0) = Z(x,1) = 1\, ,
\end{gather}
which can be derived using the principle of induction. This relation can be used to compute values of $Z(x,N)$ numerically.

\begin{figure}
\centering
\includegraphics[width=\textwidth]{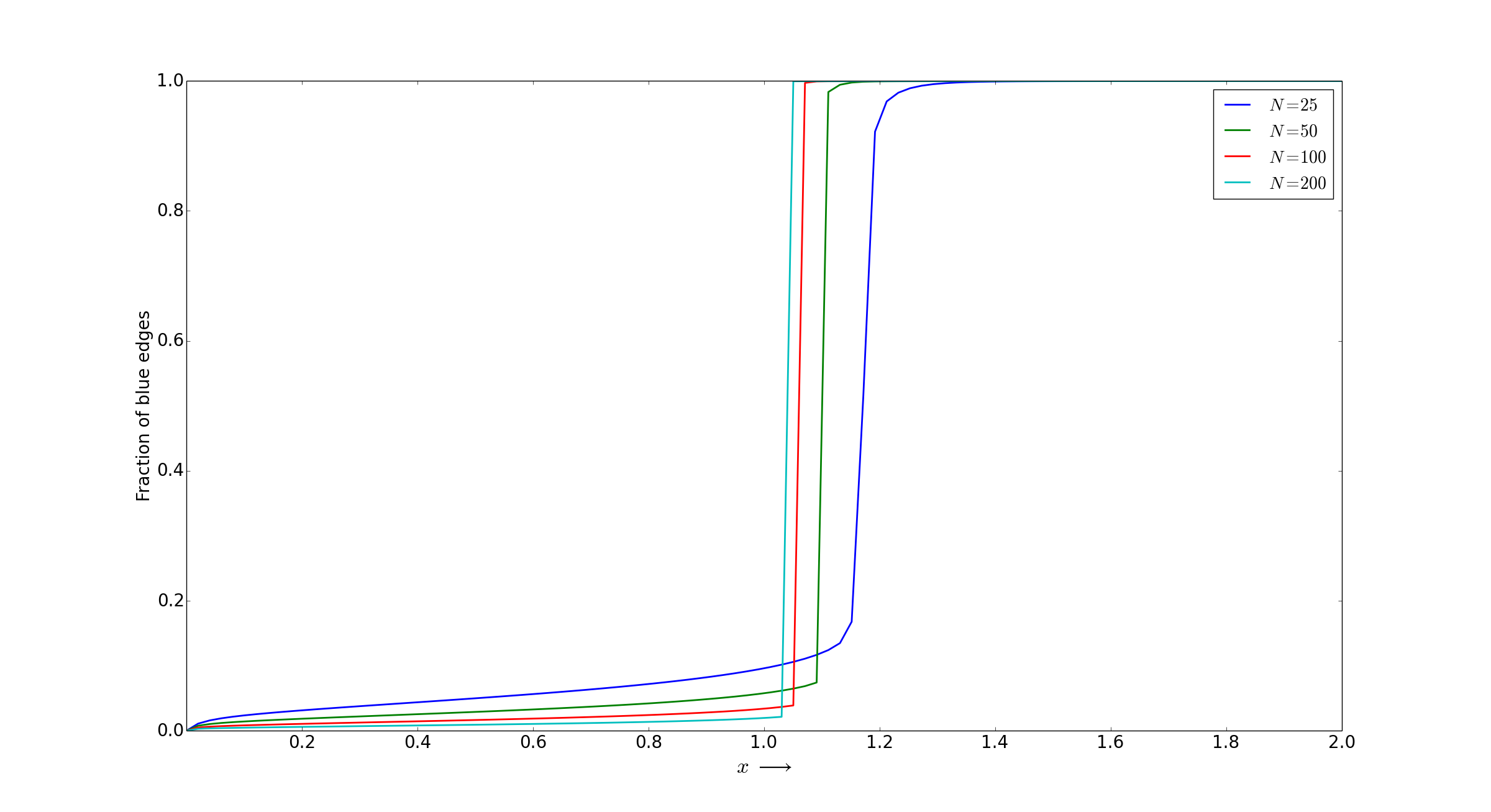}
\caption{The expected value of the fraction of blue edges averaged over the ensemble of clustering configurations depicted as a function of the parameter $x$ for various values of $N$.}
\label{fig:blue-fraction}
\end{figure}

Intuitively, the effect of the parameter $x$ is to favor or disfavor clustering configurations based on the number of blue edges. This can be quantified by calculating the number of blue edges averaged over the space of clustering configurations using the prior distribution \eqref{eq:full-prior}
\begin{eqnarray}
\langle b \rangle = x\frac{d}{dx} \log Z(x,N)\, .
\end{eqnarray}
The behavior of the blue edge fraction is shown in Figure \ref{fig:blue-fraction}.
We see a phase transition, which in the $N\rightarrow \infty$ limit is a discontinuity at $x=1$. 
Phase transitions of this sort occur in the large $N$ limit and arise when there is a balance between entropic and energetic considerations. When $x = 1 + \epsilon$, $\epsilon > 0$, there is an exponentially larger weight associated with configurations with more blue edges.
In the large $N$ limit, $\log Z(1+\epsilon,N) = N(N-1)/2 \log(1+\epsilon)$, which is the contribution to the partition function from the configuration with all blue edges. This is the ordered phase. We estimate the entropy associated with the number of clustering configurations in the large $N$ limit as $\log Z(1,N) \sim N \log N$. The balance gives us an estimate of the location of the phase transition as 
\begin{eqnarray}
x_{critical} = 1 + \epsilon, \quad \epsilon \sim \frac{2}{N} \log N\, .
\end{eqnarray}

The family of priors in \eqref{eq:full-prior} imposes a non-uniform prior on the number of clusters. To calculate expectation values we add another parameter $\lambda$ to the partition function
\begin{eqnarray}
Z_\lambda(x,N) = \sum_{C} x^{b(C)} \lambda ^{n(C)}\, ,
\end{eqnarray}
where $n(C)$ is the number of clusters, and the sum is over all clustering configurations. Clearly, $Z_{\lambda =1}(x,N) = Z(x,N)$, and taking derivatives with respect to $\lambda$ allows us to calculate moments
\begin{eqnarray}
\mu_n & = & <n> = \left. \frac{d}{d\lambda} Z_\lambda(x,N) \right|_{\lambda = 1} \, ,\\
\sigma_n & = & <n^2> - <n>^2 = \left. \left(\lambda \frac{d}{d\lambda} \right)^2 Z_\lambda(x,N) \right|_{\lambda = 1}\, .
\end{eqnarray}
The function $Z_\lambda(x,N)$ can be calculated using the recurrence relation
\begin{gather}
Z_\lambda (x,N+1) = \lambda \sum_{k=0}^N \binom{N}{k}\ x^{k(k+1)/2}\ Z_\lambda(x,N-k)\, , \\
 Z_\lambda(x,0) = 1, \  Z_\lambda (x,1) = \lambda\, .
\end{gather}

\begin{figure}
\includegraphics[width=\textwidth]{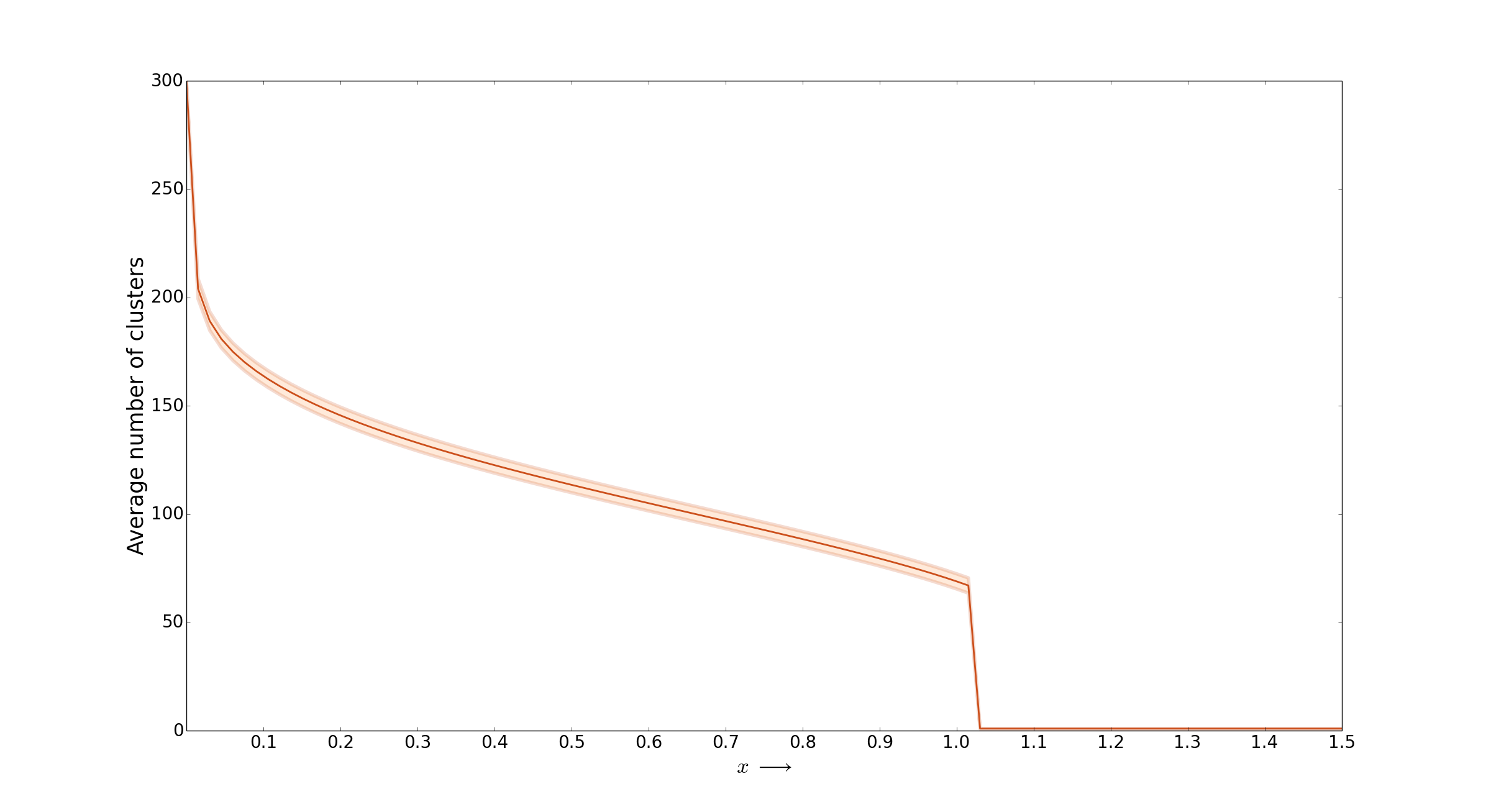}
\caption{The line shows the mean number of clusters as a function of $x$. The shading indicates one standard deviation on either side. The plot is for $N=300$ points.}
\label{fig:prior-clusters}
\end{figure}

The prior distribution is peaked over configurations with a definite number of clusters as shown in Figure \ref{fig:prior-clusters}.

We note that the parameter $x$ can be tuned by the user between 0 and 1 in order to influence the outcome of the clustering based on prior knowledge either about the number of clusters or the fraction of blue edges. We recommend the uniform prior corresponding to the choice $x=1$ that weighs all clustering configurations equally.

\section{Simplifying the message update equations}
\label{sec:simplify-update}

We recall here the message update equations from section \ref{sec:message-passing}.

\begin{eqnarray}
\outmessage_{ij\rightarrow ijk}(H_{ij}) & = & S_{ij}(H_{ij}) + \sum_{l \neq i, 
j, k} \alpha_{ij \leftarrow ijl}(H_{ij})\, , \\
\alpha_{ij \leftarrow ijk}(h_{ij}) & = & \max_{H_{jk}, H_{ki}} \big[ 
\delta_{ijk}(H_{ij}, H_{jk}, H_{ki}) + \outmessage_{jk\rightarrow ijk}(H_{jk}) 
+ \outmessage_{ki \rightarrow ijk}(H_{ki})\big]\, .
\end{eqnarray}

\subsubsection*{Simplification 1: eliminate $\outmessage$}
Since the messages $\outmessage_{ij \rightarrow ijk}$ play no role in 
determining the solution $H^*$, they can be eliminated giving a single update 
for the message $\alpha_{ij \leftarrow ijk}$, given below
\begin{eqnarray}
\alpha_{ij \leftarrow ijk}(H_{ij}) = \max_{H_{jk}, H_{ki}} \left( \delta_{ijk} +
S_{jk}(H_{jk}) + \sum_{l \neq i, j, k} \alpha_{jk \leftarrow jkl}(H_{jk}) 
+ S_{ki}(H_{ki}) + \sum_{l \neq i, j, k} \alpha_{ki \leftarrow ikl}(H_{ki}) 
\right)\,. \label{eq:alpha-update}
\end{eqnarray}

\subsubsection*{Simplification 2: Only the difference of messages matters}
Equation \eqref{eq:best-hypothesis} can be rewritten as
\begin{eqnarray}
H^*_{ij} & = & \mathrm{Step}\left(S_{ij}(1) - S_{ij}(0) + \sum_{k \neq i,j} 
[\alpha_{ij \leftarrow ijk}(1) - \alpha_{ij \leftarrow ijk}(0) ] \right)\, , \\
\mathrm{Step}(x) & = & \left\{\begin{array}{cl}
1 & \mbox{ if } x > 0 \\
0 & \mbox{ otherwise.}
\end{array}
\right. \,
\end{eqnarray}
Note that $H^*$ is only dependent on the differences
\begin{eqnarray}
\Delta S_{ij}  :=  S_{ij}(1) - S_{ij}(0)\, ,\quad 
A_{ijk}   :=  \alpha_{ij \leftarrow ijk}(1) - \alpha_{ij 
\leftarrow ijk}(0)\, .
\end{eqnarray}
Moreover, as we shall see shortly, the update for the difference $A_{ijk}$ is 
completely determined by the value of $A_{ijk}$  alone.

Equation \eqref{eq:alpha-update} can be written explicitly as 
\begin{eqnarray}
\alpha_{ij \leftarrow ijk}(1) & = & \max_{H_{jk}, H_{ki} = (1,1), (1,0), (0,1) 
} \left( 
S_{jk}(H_{jk}) + \sum_{l \neq i, j, k} \alpha_{jk \leftarrow jkl}(H_{jk}) 
+ S_{ki}(H_{ki}) + \sum_{l \neq i, j, k} \alpha_{ki \leftarrow ikl}(H_{ki}) 
\right)\,, \nonumber  \\
& = & S_{jk}(0) + \sum_{l \neq i, j, k} \alpha_{jk \leftarrow jkl}(0) 
+ S_{ki}(0) + \sum_{l \neq i, j, k} \alpha_{ki \leftarrow ikl}(0)   \nonumber \\
& & +  \max \left\{ 0, \ \ \Delta S_{jk} + \sum_{l \neq i, j, k} A_{jkl},  \ \ \Delta S_{ki} + \sum_{l \neq i, j, k} A_{kil} \right\}  \label{eq:alpha_0}\, .
\end{eqnarray}
In the first step the $\delta_{ijk}$ either contributes $-\infty$ or nothing at 
all. Since the $-\infty$ contribution never wins in the $\max()$ function those 
configurations of $H_{ij}, H_{jk}, H_{ki}$ are effectively eliminated. 
Similarly, 
\begin{eqnarray}
\alpha_{ij \leftarrow ijk}(0) & = & \max_{H_{jk}, H_{ki} = (1,1), (0,0)} \left( 
S_{jk}(H_{jk}) + \sum_{l \neq i, j, k} \alpha_{jk \leftarrow jkl}(H_{jk}) 
+ S_{ki}(H_{ki}) + \sum_{l \neq i, j, k} \alpha_{ki \leftarrow ikl}(H_{ki}) 
\right)\, , \nonumber \\
& = & S_{jk}(0) + \sum_{l \neq i, j, k} \alpha_{jk \leftarrow jkl}(0) 
+ S_{ki}(0) + \sum_{l \neq i, j, k} \alpha_{ki \leftarrow ikl}(0)   \nonumber \\
& & +  \max \left\{ 0, \ \ \Delta S_{jk} + \sum_{l \neq i, j, k} A_{jkl}+  
\Delta S_{ki} + \sum_{l \neq i, j, k}A_{kil} \right\} \, . \label{eq:alpha_1}
\end{eqnarray}
Taking the difference of equations \eqref{eq:alpha_1}, \eqref{eq:alpha_0}, we 
arrive at an update for the $A_{ijk}$:
\begin{eqnarray}
\begin{array}{ccl}
A_{ijk} & \leftarrow & \max \left\{ 0, \ \ \Delta S_{jk} + \sum_{l \neq i, j, 
k} A_{jkl}+  \Delta S_{ki} + \sum_{l \neq i, j, k} A_{kil} \right\} - 
\nonumber \\
& &  \max \left\{ 0, \ \ \Delta S_{jk} + \sum_{l \neq i, j, k} A_{jkl},  \ \ 
\Delta S_{ki} + \sum_{l \neq i, j, k} A_{kil} \right\}
\end{array}\, .
\end{eqnarray}
In each iteration of the algorithm we have to update $\cO(N^3)$ variables $A_{ijk}$, each of which involves a sum over $\cO(N)$ terms. This makes the complexity $\cO(N^4)$. The run time scaling with $N$ can be improved to $\cO(N^3)$ by computing the summations ahead of time. We introduce the matrix $B_{ij} := \Delta S_{ij} + \sum_{k \neq i,j} A_{ij}(k) $ in terms of which the update to the $A_{ijk}$ becomes
\begin{eqnarray}
A_{ijk} \leftarrow \max \left\{ 0, \ \ B_{jk} - A_{jki}+  B_{ki}  -  A_{kij} 
 \right\} 
	   -  \max \left\{ 0, \ \ B_{jk} - A_{jki},  \ \ B_{ki} - A_{kij} \right\} \, ,
\end{eqnarray}
and the best configuration is obtained by
\begin{equation}
H^*_{ij}  = \mathrm{Step}(B_{ij})\, .
\end{equation}

\end{document}